\documentclass[letterpaper, 10 pt, conference]{ieeeconf}

\IEEEoverridecommandlockouts
\usepackage{amsmath,amsfonts}
\usepackage{algorithmic}
\usepackage{algorithm}
\usepackage{multirow}
\usepackage{xcolor}
\usepackage{makecell}
\usepackage{array}

\usepackage{textcomp}
\usepackage{stfloats}
\usepackage{url}
\usepackage{verbatim}
\usepackage{graphicx}
\usepackage{tabularx}
\usepackage{cite}
\usepackage{optidef}
\usepackage{subfig} 
\usepackage{booktabs}    
\usepackage[colorlinks, linkcolor=red, anchorcolor=green, citecolor=blue ]{hyperref} 
\newenvironment{IEEEkeywords}{}{}

\title{\LARGE \bf
Estimating Continuum Robot Shape under External Loading using Spatiotemporal Neural Networks}

\author{Enyi Wang$^{1,2}$, Zhen Deng$^{2,*}$, Chuanchuan Pan$^{2}$, 
Bingwei He$^{2}$, Jianwei Zhang$^{3}$%
\thanks{This work was partly supported by National Key R\&D Program of China (Grant 2024YFB4710200) and by the Fujian Provincial Department of Science and Technology (Grant 2024I0005, Grant 2023J011175, Grant 2024Y0011 and Grant 2024H0027). (Corresponding author: Zhen Deng, zdeng@fzu.edu.cn).}%
\thanks{$^{1}$ Hamlyn Centre for Robotic Surgery, Institute of Global Health Innovation, Imperial College London, SW7 2AZ, UK.}%
\thanks{$^{2}$ Department of Mechanical Engineering and Automation, Fuzhou University, Fuzhou 350108, China.}%
\thanks{$^{3}$ TAMS Group, Informatics, University of Hamburg, D-22527, Hamburg, Germany.}%
}

\begin{document}
\maketitle

\begin{abstract}
This paper presents a learning-based approach for accurately estimating the 3D shape of flexible continuum robots subjected to external loads. The proposed method introduces a spatiotemporal neural network architecture that fuses multi-modal inputs, including current and historical tendon displacement data and RGB images, to generate point clouds representing the robot's deformed configuration. The network integrates a recurrent neural module for temporal feature extraction, an encoding module for spatial feature extraction, and a multi-modal fusion module to combine spatial features extracted from visual data with temporal dependencies from historical actuator inputs. Continuous 3D shape reconstruction is achieved by fitting Bézier curves to the predicted point clouds. Experimental validation demonstrates that our approach achieves high precision, with mean shape estimation errors of 0.08 mm (unloaded) and 0.22 mm (loaded), outperforming state-of-the-art methods in shape sensing for TDCRs. The results validate the efficacy of deep learning-based spatiotemporal data fusion for precise shape estimation under loading conditions.

\end{abstract}


\section{Introduction}
Tendon-driven continuum robots (TDCRs) have gained prominence due to their remarkable flexibility and compliance, enabling safe navigation through confined environments~\cite{russo2023continuum,nguyen2015tendon}. Unlike conventional rigid robots, TDCRs undergo continuous deformation without inducing  damage~\cite{kolachalama2020continuum,dupont2022continuum,vitiello2012emerging}. Their motion is actuated by tendon-driven mechanisms, which generate bending or twisting through controlled tendon tension. External loadings, such as interactive forces or gravitational effects, alter the deformation behavior of TDCRs, thereby complicating the prediction of 3D shapes. Accurate 3D shape estimation of TDCRs is critical for their control, enabling TDCRs to dynamically adapt their configuration for optimal task execution and environmental interaction. However, achieving precise shape estimation under loading conditions remains challenging due to mechanical hysteresis, material nonlinearity, and sensor limitations~\cite{shi2016shape}.
\begin{figure}
    \centering
    \includegraphics[width=\columnwidth]{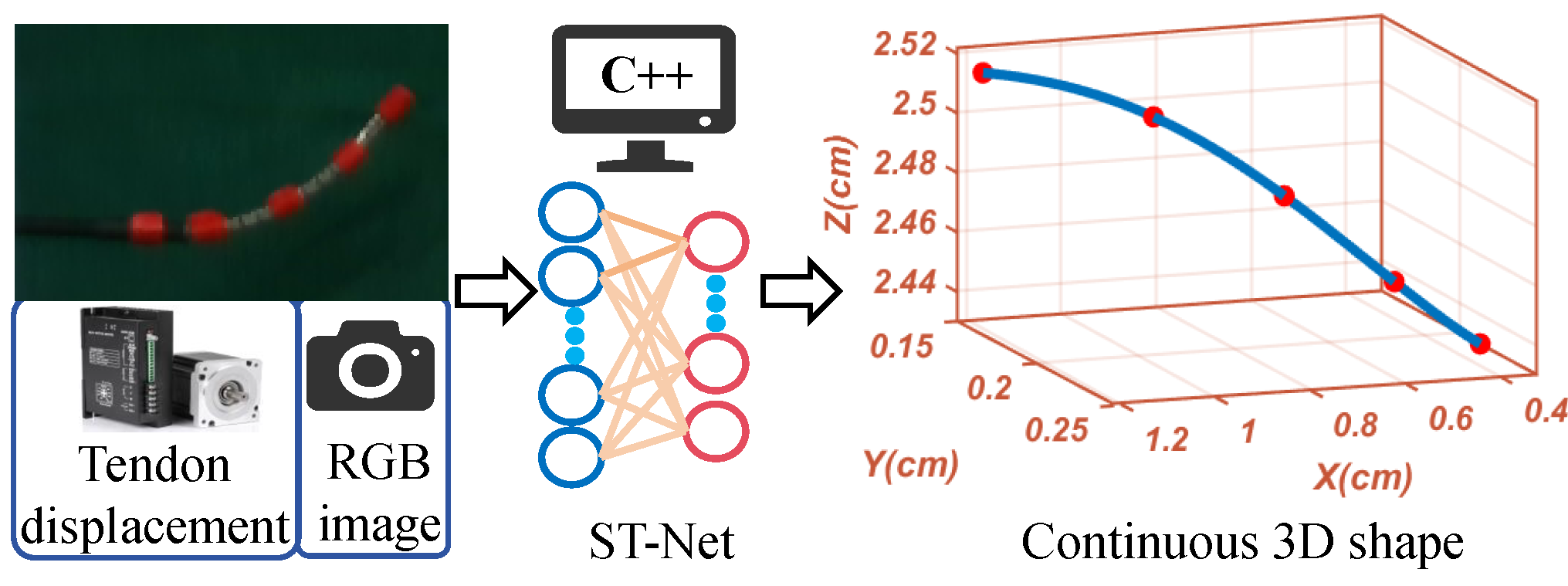}
    \caption{Schematic of the proposed learning-based framework for 3D shape estimation of TDCRs under external loading.}
    \label{fig:overall}
\end{figure}

Various sensors have been used for the shape detection of TDCRs, including electromagnetic (EM) sensors~\cite{song2015electromagnetic}, strain sensors~\cite{monet2020high}, and vision sensors~\cite{vandini2015vision}. Song et al.~\cite{song2015electromagnetic} proposed an EM sensor-based shape-sensing method in flexible robots. However, EM sensors provide discrete pose data for TDCRs, which limits reconstruction accuracy. Fiber Bragg Grating (FBG) sensors enable continuous strain monitoring along the robot’s backbone, facilitating shape reconstruction through strain integration. In~\cite{rahman2019modular}, a modular FBG bending sensor was developed to measure the deflection of a continuum neurosurgical robot. Nevertheless, optical fibers in FBG sensors are fragile and prone to damage under significant bending, restricting their practical utility. Vision-based approaches offer cost-effective and non-invasive alternatives. Shentu et al.~\cite{shentu2023moss} developed a deep neural network to process RGB images and output point clouds representing the robot's shape. However, such methods are susceptible to occlusions and illumination variability. Multi-view camera systems can mitigate occlusion issues~\cite{shi2016shape}, but they impose system complexity.

Model-based methods have been proposed to estimate the shape of TDCRs using actuation inputs~\cite{webster2010design}. Kinematic models, including constant curvature~\cite{rao2022shape}, Cosserat-rod~\cite{chen2024chained}, and Finite Element methods~\cite{guo2019preliminary}, have been developed to describe the deformation of TDCRs. However, such physics-based approaches often fail to account for material nonlinearity, unknown external loads, and dynamic effects such as friction and hysteresis, resulting in discrepancies between predicted and observed shapes. Recent advancements in data-driven methods have enabled nonlinear mappings between actuation inputs and the shape of TDCRs without explicit physical modeling~\cite{chen2024data}. In~\cite{ha2022shape}, a neural network was trained to estimate shape from FBG wavelength shifts, whereas Hao et al.~\cite{hao2023two} mapped proximal tendon displacements to 3D point clouds via a multilayer perception, which outlined the shape of TDCRs. Kuntz et al.~\cite{kuntz2020learning} developed a feed-forward neural network for a concentric tube robot, using tube rotations and translations as inputs to predict the robot's shape. Despite their potential, these methods exhibit limited robustness under unknown external loads. To avoid this, Zhao et al.~\cite{zhao2021shape} introduced an LSTM-based neural network that predicts point clouds along the robot's backbone by combining strain gauge readings and actuation inputs, enabling shape prediction under external loading. Tanaka et al.~\cite{tanaka2022continuum} combined actuation input and proprioceptive bending sensor outputs for shape estimation using recurrent neural networks. Incorporating multimodal sensor data as input to the neural network can enhance the precision of robot shape estimation. However, these approaches heavily rely on specific sensor-actuator mapping scenarios.
\begin{figure*}
    \centering
    \includegraphics[width=\textwidth]{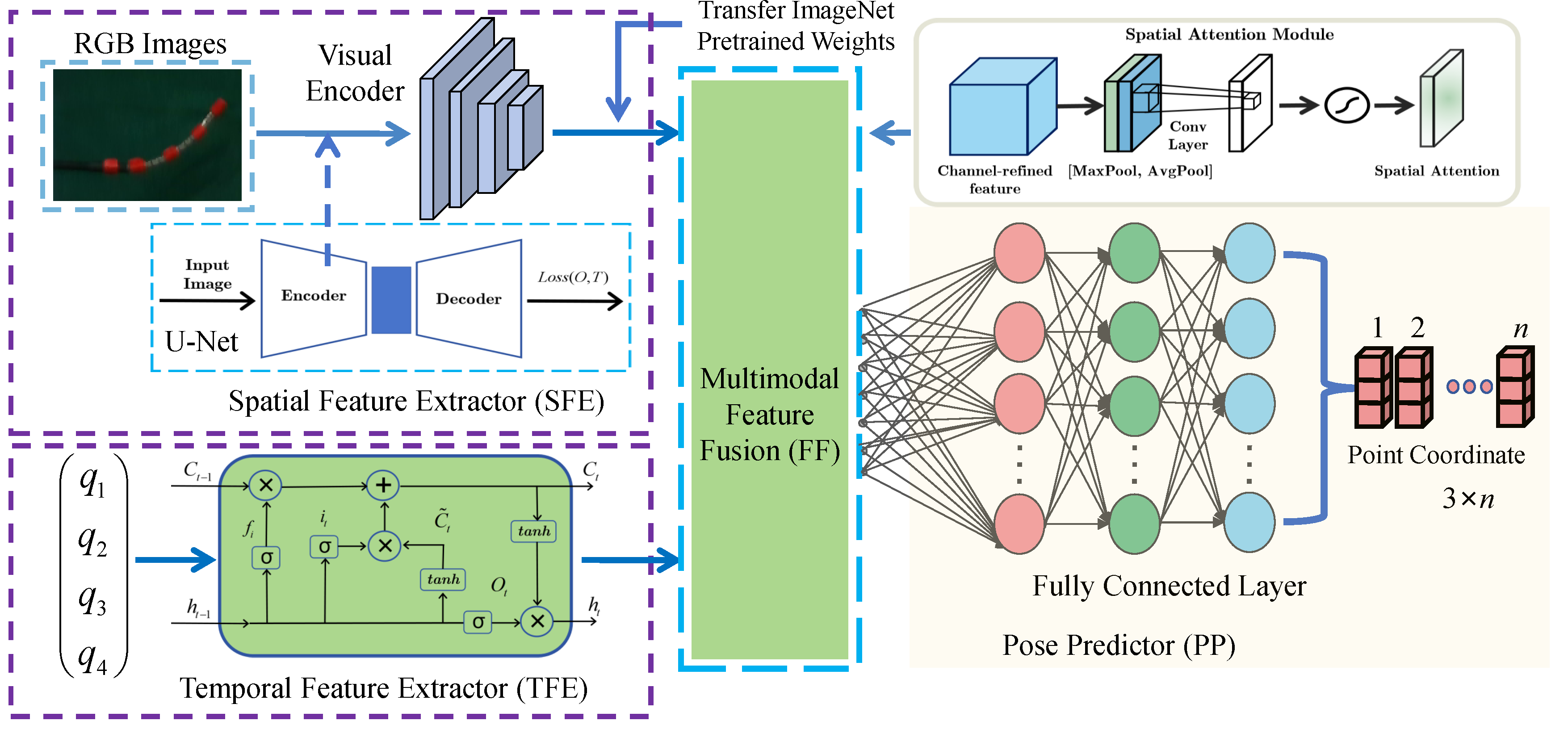}
    \caption{Structural overview of the proposed spatial-temporal neural network (ST-Net) combining visual perception and proprioceptive sensing.}
    \label{ST-Net}
\end{figure*}

This study proposes a learning-based framework for accurate 3D shape estimation of TDCRs under external loading. A spatiotemporal neural network is developed to model the nonlinear relationship between the robot’s shape and multimodal sensor data, consisting of tendon displacement data and RGB images. The network extracts high-dimensional spatial features from images and low-dimensional temporal features from tendon displacement data, which are dynamically fused through a soft attention mechanism to prioritize salient deformation cues. Continuous 3D shape reconstruction is achieved by fitting Bézier curves to the point clouds predicted by the network. This approach eliminates dependence on analytical mechanical models while accommodating external loads. Extensive experiments validate the effectiveness of the approach, including comparative evaluations against state-of-the-art methods. Results demonstrate that our framework achieves accurate shape estimation of TDCRs under varying external loads.

\section{Methodology}
This section introduces the proposed learning-based shape estimation framework for TDCRs using a spatial-temporal neural network. As illustrated in Fig.~\ref{fig:overall}, the TDCR is actuated by four tendons, with their displacements parameterized as $\mathbf{q} = [q_1, q_2, q_3, q_4]^{T} \in \mathbb{R}^4$. The shape estimation task aims to reconstruct the point clouds $\mathbf{P} = [p_1, p_2, \ldots, p_n]^{T} \in \mathbb{R}^{n \times 3}$ along the robot's backbone. A monocular RGB camera is used to capture robot images $\mathbf{I}$. The proposed spatiotemporal neural network (ST-Net) takes $\mathbf{q}$ and $\mathbf{I}$ as inputs to predict $\mathbf{P}$. Then, the continuous 3D shape of TDCRs is reconstructed by fitting a parametric Bézier curve to $\mathbf{P}$. The following subsections detail the network architecture, loss function design, and shape reconstruction methods.

\subsection{Network Architecture}
The ST-Net architecture consists of four main modules: the Spatial Feature Extractor (SFE), Temporal Feature Extractor (TFE), Feature Fusion (FF), and Pose Predictor (PP). As illustrated in Fig.~\ref{ST-Net}, the SFE extracts spatial features from input RGB images, while the TFE processes temporal tendon displacement sequences. The FF module integrates these multimodal features, which are fed into the PP module to predict 3D coordinates of point clouds representing the robot's shape.
 
\textit{SFE module}: The SFE module adopts an encoder architecture inspired by U-Net~\cite{ronneberger2015u} for visual feature extraction from RGB images. The encoder consists of four consecutive convolutional blocks, each containing two $3 \times 3$ convolutional layers with Rectified Linear Unit (ReLU) activations~\cite{glorot2011deep}, followed by a $2 \times 2$ max-pooling layer. This architecture progressively reduces spatial resolution while expanding channel dimensions from $3$ to $1024$. The output of the encoder is the extracted spatial features from the input images. The SFE module is pretrained via unsupervised learning with an image reconstruction loss function.

\textit{TFE module}: Temporal dependencies in tendon displacement vectors $\mathbf{q}$ are captured using a Long Short-Term Memory (LSTM) network~\cite{shi2015convolutional}. The LSTM maintains an internal cell state $c_t$ and hidden state $h_t$ through gating mechanisms that regulate information flow:
\begin{equation}
    \begin{gathered}
    f_t = \sigma_g\left(W_f \cdot\left[h_{t-1}, q_t\right] + b_f\right) \\
    i_t = \sigma_g\left(W_i \cdot\left[h_{t-1}, q_t\right] + b_i\right) \\
    o_t= \sigma_g\left(W_o \cdot\left[h_{t-1}, q_t\right]+b_o\right) \\
    \tilde{c}_t = \sigma_c \left(W_c \cdot\left[h_{t-1}, q_t\right] + b_c\right) \\
    c_t= f_t \odot c_{t-1}+i_t \odot \tilde{c}_t \\    
    h_t= o_t \odot \sigma_c \left(c_t\right)
    \end{gathered}
    \label{eq1}
\end{equation} 
where $f_t$, $i_t$, and $o_t$ represent forget, input, and output gates, respectively. $\sigma_g$ and $\sigma_c$ denote the sigmoid and hyperbolic tangent activation, respectively, while $\odot$ indicates element-wise multiplication. This architecture enables effective modeling of temporal patterns in actuation sequences.

\textit{FF module}: The fusion of Spatial features from images and temporal features from tendon input is achieved through a spatial attention mechanism. Given concatenated features $\mathbf{F} \in \mathbb{R}^{H \times W \times C}$ from spatial and temporal streams, where $H$, $W$, and $C$ denote height, width, and channels, respectively. The attention map $\mathbf{M} \in \mathbb{R}^{H \times W \times 1}$ is computed as:
\begin{equation} \mathbf{M} = \sigma_g \Big(\text{Conv}_{7\times7}\big(\text{Concat}(\mathbf{F}{\text{avg}}, \mathbf{F}_{max})\big)\Big) \in \mathbb{R}^{H \times W \times 1}  \end{equation}
with 
$$\mathbf{F}{\text{avg}} = \text{AvgPool}(\mathbf{F}) \in \mathbb{R}^{H \times W \times 1}$$
$$\mathbf{F}{\text{max}} = \text{MaxPool}(\mathbf{F}) \in \mathbb{R}^{H \times W \times 1}$$
where $\text{Conv}_{7\times7}$ denotes a 2D convolutional with $7 \times 7$ kernels. The final refined features are obtained via $\mathbf{F}' = \mathbf{M} \cdot \mathbf{F}$, emphasizing discriminative spatial regions through attention-weighted fusion. 

\textit{PP module}: The final module processes the fused features $\mathbf{F}'$  through a dropout layer (rate $p=0.5$) for regularization, followed by a fully connected layer that projects the features to a $3 \times n$-dimensional output space:
\begin{equation} \mathbf{P} = \mathbf{W}{fc}\cdot\text{Dropout}(\mathbf{F}') + \mathbf{b}{fc} \in \mathbb{R}^9 \end{equation} 
where $\mathbf{P}$ represents the 3D coordinates for point clouds outlining the TDCR's shape configuration. This representation balances reconstruction accuracy with computational efficiency.

\subsection{Loss Function Design}
The SFE undergoes unsupervised pretraining using a composite loss function $l_{\text{SFE}}$, comprising three components: Mean Squared Error loss ($l^{m}_{\text{SFE}}$), Perceptual Loss ($l^{p}_{\text{SFE}}$), and Structural Similarity Index Loss ($l^{s}_{\text{SFE}}$). These components collectively enhance spatial feature extraction through pixel-level accuracy, semantic consistency, and structural preservation. The composite loss $l_{\text{SFE}}$ is formulated as:
\begin{equation}
{l_{\text{SFE}}} = {l^{m}_{\text{SFE}}} + \alpha \cdot {l^{p}_{\text{SFE}}} + \beta \cdot {l^{s}_{\text{SFE}}}
\label{eq3}
\end{equation}

\textit{Mean Squared Error Loss ($l^{m}_{\text{SFE}}$)}: computes pixel-wise intensity differences between predicted and target image:
\begin{equation}
{l^{m}_{\text{SFE}}} = \frac{1}{n}\sum\limits_{i = 1}^n {{{\left( {I_{p}^{(i)} - I_{t}^{(i)}} \right)}^2}}
\label{eq4}
\end{equation}
where $n$ is the total number of pixels.

\textit{Perceptual Loss ($l^{p}_{\text{SFE}}$)}: measures feature-space discrepancies using a pre-trained network $\phi $, comparing high-level representations from $N$ selected layers:
\begin{equation}
{l^{p}_{\text{SFE}}} = \sum_{i=1}^N {\left| \phi_i(I_p) - \phi_i(I_t) \right|_2^2}
\label{eq5}
\end{equation}
where ${\phi_i}$ denotes feature maps from the $i-th$ network layer, and $\left| \cdot \right|_2^2$ denotes the squared L2-norm.

\textit{Structural Similarity Index Loss ($l^{s}_{\text{SFE}}$)}: evaluates structural similarity between predicted and target images through luminance, contrast, and texture comparisons.
\begin{equation}
\begin{aligned}
SSIM\left( {{I_{p}},{I_{t}}} \right) &= \frac{{\left( {2{\mu_{{I_{p}}}}{\mu_{{I_{t}}}} + {c_1}} \right)\left( {2{\sigma_{{I_{p}}{I_{t}}}} + {c_2}} \right)}}{{\left( {\mu_{{I_{p}}}^2 + \mu_{{I_{t}}}^2 + {c_1}} \right)\left( {\sigma_{{I_{p}}}^2 + \sigma_{{I_{t}}}^2 + {c_2}} \right)}}\\
{l^{s}_{\text{SFE}}} &= 1 - SSIM\left( {{I{p}},{I_{t}}} \right)
\end{aligned}
\label{eq6}
\end{equation}
where $\mu_{{I_{p}}}$ and $\mu_{{I_{t}}}$ denote mean intensities of $I_{p}$ and $I_{t}$, respectively. $\sigma_{{I_{p}}}^2$ and $\sigma_{{I_{t}}}^2$ represent intensity variances, and $\sigma_{{I_{p}}{I_{t}}}$ is the sample covariance. Constants $c_1$ and $c_2$ prevent division by zero. 

For shape prediction, the ST-Net is trained using a Mean Squared Error loss between predicted and ground-truth 3D point clouds:
\begin{equation}
L_{\text{ST}}  =\frac{1}{N} \sum_{i=1}^N \left( \mathbf{P}_{p} - \mathbf{P}_{t} \right)^2
\label{eq7}
\end{equation}
where $\mathbf{P}_{p},\mathbf{P}_{t} \in \mathbb{R}^{n\times3}$ denote predicted and ground-truth point clouds respectively, with $n$ points defining the robot's shape.

\subsection{Continuous 3D Shape Reconstruction}
The proposed method converts discrete point clouds $\mathbf{P}$  predicted by the ST-Net into a continuous 3D shape via Bézier curve fitting. A $4$th-order Bézier curve was employed for this reconstruction, requiring the determination of optimal control points to approximate the predicted point clouds. The initial coordinates of the robot base, which remained unchanging, were directly incorporated into the fitting process, defined as:
\begin{equation} 
B(t) = \sum\limits_{i = 0}^4 \binom{4}{i} p_i(1 - t)^{4 - i} t^i, \quad t \in [0,1] 
\label{eq8} 
\end{equation} 
where $p_i$ represents the $i$-th control point.

Control points of the Bézier curve are optimized to minimize the geometric error between predicted points and curve positions. The least-squares objective function is formulated as:
\begin{equation} 
J = \sum_{k=1}^N \left( \mathbf{P}_k - B(t_k) \right)^2 
\end{equation}
where $\mathbf{P}_k$ and $B(t_k)$ denote the $k$-th predicted point and curve position, and $t_k$ is computed via normalized arc-length parameterization from the fixed base.

The optimization process adjusted all control points except the fixed base point, generating a smooth curve that accurately reconstructs the TDCR's shape configuration. This approach ensures continuity along the reconstructed shape while maintaining computational efficiency through the convex hull property of Bézier curves.

\section{Experiments}
\begin{figure}[htbp!]
    \centering
    \includegraphics[width=\columnwidth]{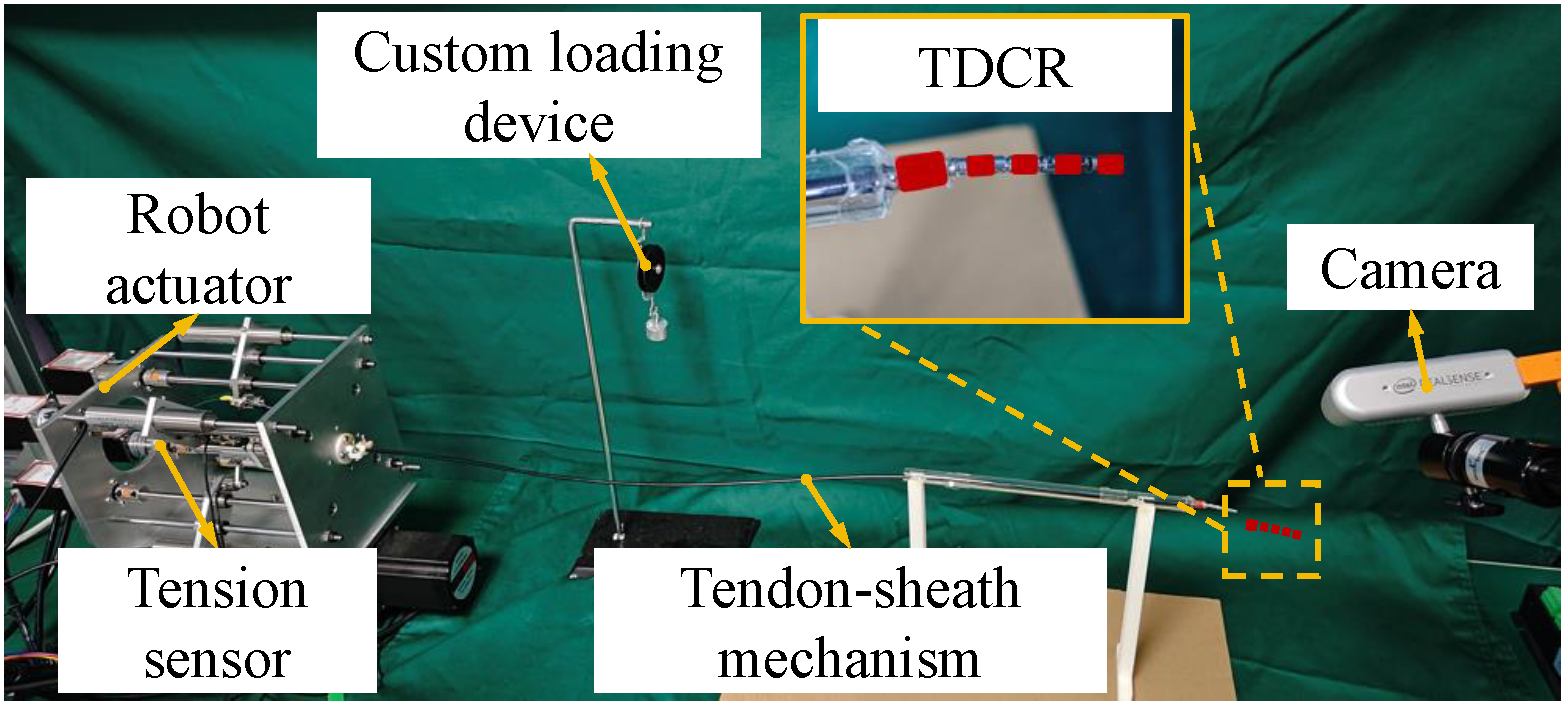}
    \caption{Experimental setup for shape estimation validation.}
    \label{fig:setup}
\end{figure}

Experimental validation was conducted using a custom-designed 2-degree-of-freedom TDCR driven by four independent tendon actuators. As shown in Fig.~\ref{fig:setup}, the experimental setup incorporates an RGB-D camera (Intel RealSense D435) to capture 3D marker coordinates and a custom loading device that applies external forces through weighted pulleys, simulating realistic load-constrained scenarios.

\subsection{Datasets Construction and Model Training}
A dataset was generated by systematically adjusting the tendon displacements using controlled motor commands, inducing bending angles within $[-60^\circ, 60^\circ]$ in multiple directions. The RGB-D camera acquired $640 \times 480$ resolution RGB images, with subsequent point cloud processing extracting 3D coordinates from five fiducial markers on the robot surface. The ground truth coordinates were computed by projecting detected 2D marker centroids into 3D space using corresponding depth values. Five motion cycles were executed to produce $880$ samples per trial, each containing: tendon displacement measurements, ground-truth 3D marker coordinates, and time-synchronized RGB images. The dataset was partitioned into training and testing sets using an $8:2$ ratio.

The network architecture employs a two-layer LSTM in the temporal feature extraction module with hidden dimensions $100$, optimized using Adam with a learning rate of $0.001$ and batch size $64$. We set loss weights $\alpha$ and $\beta$ to $0.5$ through empirical validation. All implementations used PyTorch 1.8.1 on an NVIDIA RTX 3090 GPU, with the final shape estimation system run on an Intel Core i9-10900K CPU through Python-based deployment.

\begin{center}
    \captionof{table}{Performance Comparison of Different Networks}
    \label{Table1}
    \small
    \setlength{\tabcolsep}{3pt}
    \begin{tabular}{@{}l *{4}{cc}@{}}
    \toprule
    \multirow{2}{*}{Segment} & 
    \multicolumn{2}{c}{\makecell{TFE-Net\\($\mathbf{q}$)}} & 
    \multicolumn{2}{c}{\makecell{SFE-Net\\($\mathbf{I}$)}} & 
    \multicolumn{2}{c}{\makecell{FF-Net\\($\mathbf{q,I}$)}} & 
    \multicolumn{2}{c}{\makecell{Proposed\\Network}} \\
    \cmidrule(lr){2-3} \cmidrule(lr){4-5} \cmidrule(lr){6-7} \cmidrule(l){8-9}
    mm & RMSE & Max & RMSE & Max & RMSE & Max & RMSE & Max \\ 
    \midrule
    $x_3$ & 0.04 & 0.12 & 0.05 & 0.20 & 0.03 & 0.09 & 0.03 & 0.07 \\
    $y_3$ & 0.10 & 0.25 & 0.48 & 1.30 & 0.05 & 0.15 & 0.04 & 0.12 \\
    $z_3$ & 0.05 & 0.13 & 0.08 & 0.23 & 0.05 & 0.14 & 0.05 & 0.13 \\
    $x_4$ & 0.18 & 0.44 & 0.23 & 0.72 & 0.08 & 0.25 & 0.05 & 0.11 \\
    $y_4$ & 0.30 & 0.61 & 1.18 & 3.10 & 0.14 & 0.42 & 0.08 & 0.24 \\
    $z_4$ & 0.08 & 0.21 & 0.15 & 0.36 & 0.05 & 0.16 & 0.04 & 0.12 \\
    $x_5$ & 0.45 & 0.99 & 0.59 & 1.42 & 0.20 & 0.58 & 0.10 & 0.33 \\
    $y_5$ & 0.54 & 1.02 & 1.86 & 4.70 & 0.27 & 0.73 & 0.17 & 0.71 \\
    $z_5$ & 0.09 & 0.27 & 0.11 & 0.32 & 0.05 & 0.24 & 0.04 & 0.13 \\
    \midrule
    Overall & 0.269 & 1.02 & 0.78 & 4.70 & 0.13 & 0.73 & \textbf{0.08} & \textbf{0.71} \\
    \bottomrule
    \end{tabular}
\end{center}

\begin{center}
    \includegraphics[width=\columnwidth]{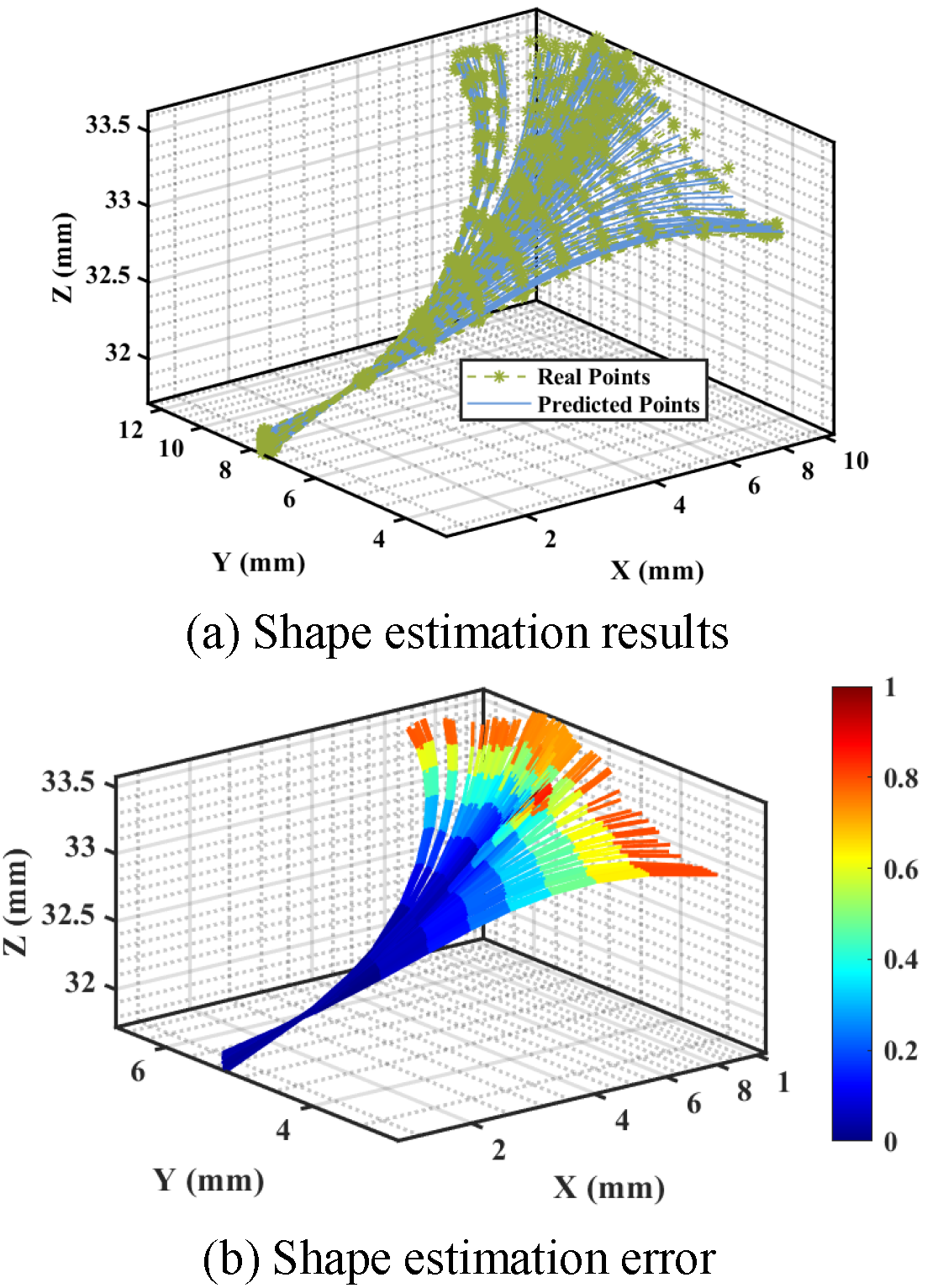}
    \captionof{figure}{Shape estimation visualization: (a) Reconstructed shapes across configurations; (b) Euclidean error heatmaps.}
    \label{fig:visualization}
\end{center}

\subsection{Ablation Study}
To evaluate the contributions of the proposed network modules in ST-Net, an ablation study was conducted with four configurations under free-space: (1) TFE-Net: A proprioceptive-only approach utilizing tendon displacements via the TFE and PP modules; (2) SFE-Net: A vision-only baseline estimating TDCR shape through sequential integration of the SFE and PP modules; (3) FF-Net: A multimodal variant combining visual and tendon displacement data without spatial attention; (4) Proposed Network: Our full architecture integrating SFE, TFE, PP, and FF modules with attention mechanisms.

Table~\ref{Table1} summarizes the quantitative results, where performance is evaluated using the Root Mean Square Error (RMSE) and Maximum Error between the predicted and ground-truth 3D coordinates. The multimodal FF-Net significantly outperformed unimodal baselines with an RMSE of $0.13$ mm ($83.1\%$ improvement over SFE-Net and $50.5\%$ over TFE-Net) and maximum error of $0.73$ mm, highlighting complementary spatial-temporal fusion. Our proposed network achieved better performance with an RMSE of $0.08$ mm and maximum error of $0.71$ mm, yielding a $39.0\% $ RMSE reduction over FF-Net. The results indicate that multimodal sensor fusion provides greater shape estimation fidelity than unimodal approaches. The attention mechanisms are essential for adaptive feature fusion between visual and proprioceptive streams. 

Fig.~\ref{fig:visualization} illustrates the shape estimation results. Fig.~\ref{fig:visualization}(a) compares reconstructed shapes across configurations against ground-truth, while Fig.~\ref{fig:visualization}(b) quantifies spatial errors via heatmaps. The proposed network achieves submillimeter accuracy along most of the robot's backbone, with peak errors ($0.82$ mm) localized at the distal tip. This progressive accuracy degradation highlights the challenge of estimating highly deformable distal segments.

\subsection{Comparison Experiment}
\begin{figure*}
\centering
\includegraphics[width=\textwidth]{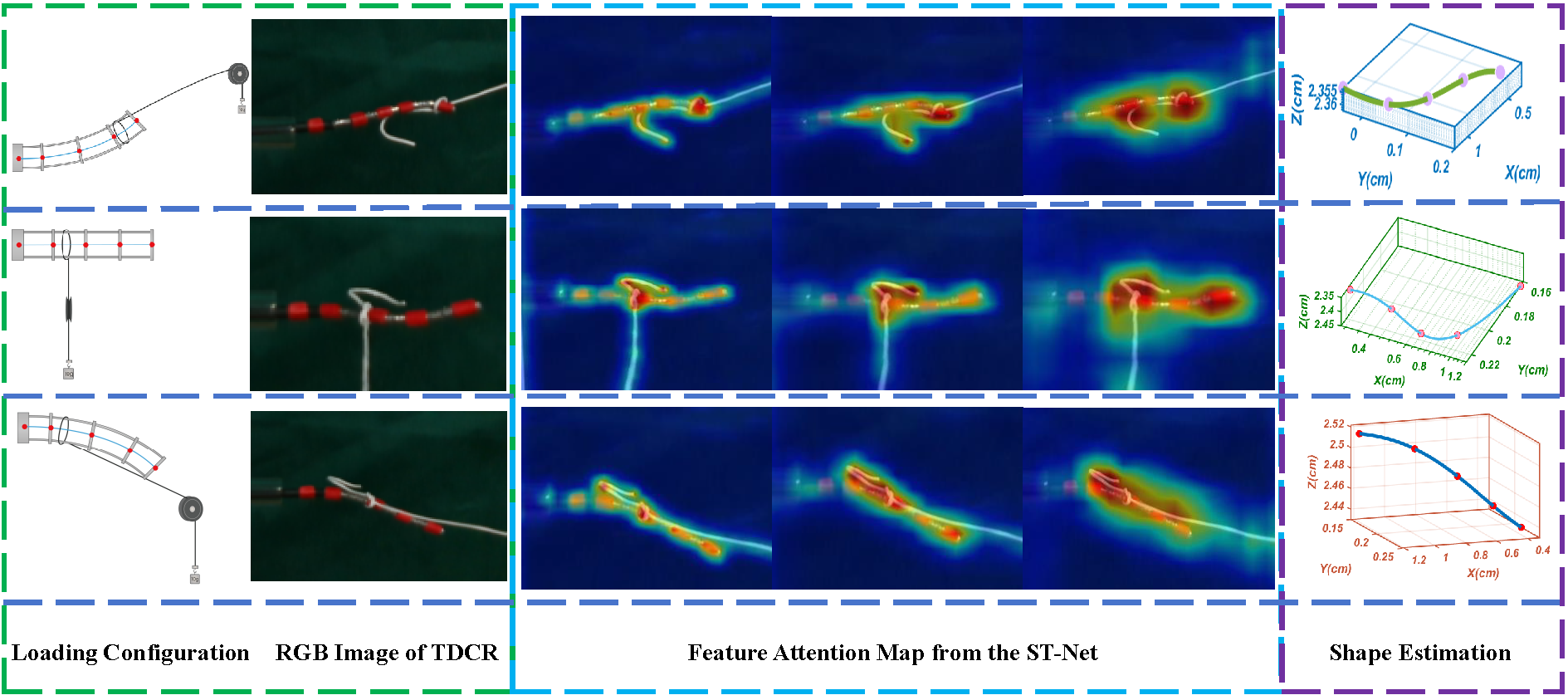}
\caption{Comparative shape estimation performance under varying external loading.}
\label{fig:estimation}
\end{figure*}

The proposed method was evaluated against two state-of-the-art methods: (1) ESN: a deep reservoir computing framework with a two-layered echo state network (ESN) developed by Tanaka et al.~\cite{tanaka2022continuum}, which estimates the shape of TDCRs from proprioceptive sensor data; (2) MoSSNet: the vision-only estimation method by Shentu et al.~\cite{shentu2023moss} that employs a deep encoder-decoder network to predict continuum robot shapes from single RGB images. Experiments were conducted under three external loading conditions ($F_{e1}, F_{e2}, \text{and} F_{e3} $) generated by a custom loading device. A $ 10g$ mass was applied through a stationary pulley system with controlled orientations to produce multidirectional loading effects.
\begin{table}[h]
\centering
\caption{Shape Estimation Performance Comparison (RMSE in mm)}
\label{Table2}
\begin{tabular}{@{}lccc@{}}
\toprule
Condition & \multicolumn{3}{c}{RMSE (mm)} \\ 
\cmidrule(l){2-4}
 & ESN & MoSSNet & Proposed method \\ 
\midrule
Free-Space & 0.31 & 0.78 & \textbf{0.08} \\
External Load $F_{e1}$ & 0.79 & 0.85 & \textbf{0.49} \\
External Load $F_{e2}$ & 0.63 & 0.82 & \textbf{0.15} \\
External Load $F_{e3}$ & 0.12 & 0.21 & \textbf{0.04} \\
\bottomrule
\end{tabular}
\end{table}

Table~\ref{Table2} compares the estimation performance across free-space and three external loading scenarios. In the free-space scenario, the proposed method achieves an RMSE of $0.08$ mm, representing $74\%$ and $89.6\%$ error reductions compared to ESN ($0.31$ mm) and MoSSNet ($0.78$ mm), respectively. Under external loading, our method outperforms both benchmarks by significant margins, with RMSE values of $0.49$ mm, $0.15$ mm, and $0.04$ mm for $F_{e1}$, $F_{e2}$, and $F_{e3} $, respectively. While ESN captures temporal relationships by leveraging a randomized reservoir layer, its sensitivity to hyperparameters (e.g., reservoir size, spectral radius, and input scaling) limits adaptability in loading conditions, leading to higher errors (e.g., $0.79$ mm at $F_{e1}$). Meanwhile, the vision-only MoSSNet lacks direct tendon-displacement cues, making it prone to inaccuracies under loading conditions. The proposed method's consistent sub-millimeter accuracy across diverse load directions confirms its robustness in loading environments.

Fig.~\ref{fig:estimation} illustrates the shape estimation results under external loading, accompanied by feature attention maps extracted from successive layers of the FF module. These attention maps demonstrate that the proposed network effectively identifies critical deformation features along the TDCR's backbone, allowing accurate spatial reconstruction despite external loading. The results highlight the advantages of multimodal data fusion for TDCR shape estimation, particularly in load-affected scenarios. This demonstrates that the proposed method achieves precise shape estimation for TDCRs under external loads.

\section{Conclusion}
This paper presents a novel learning-based approach for estimating the 3D shape of tendon-driven continuum robots (TDCRs) under external loads. The developed end-to-end neural network architecture integrates feature extraction, fusion, and pose prediction modules to correlate multimodal sensor data with the robot’s 3D configuration. By leveraging spatiotemporal input data, the network effectively combines high-dimensional spatial features from images with low-dimensional temporal features from tendon displacement, enabling accurate shape estimation even under external loads. Continuous 3D shape reconstruction is achieved through the Bézier curve, which fits the predicted point clouds from the proposed network. Experimental results demonstrate that the proposed method effectively fuses visual and tendon-displacement information, enabling accurate shape estimation for TDCRs in both unloaded and external load environments. However, this study has several limitations: the validation was conducted on a relatively short continuum robot under simple loading conditions, and the loading wire itself may act as an unintended visual cue that potentially influences the network's performance. Future work will focus on conducting extensive experiments to investigate more complex robot-environment interactions. Its applications in the shape control of TDCRs for tasks such as minimally invasive surgery will also be explored. 

\bibliographystyle{IEEEtran}
\bibliography{reference}

\end{document}